# Millimeter-Wave Sensing for Avoidance of High-Risk Ground Conditions for Mobile Robots

J. Blanche, S. Chirayil Nandakumar, D. Mitchell, S. Harper, K. Groves, A. West, B. Lennox, S. Watson, D. Flynn, I. Yamamoto

*Abstract—* Mobile robot autonomy has made significant advances in recent years, with navigation algorithms well developed and used commercially in certain well-defined environments, such as warehouses. The common link in usage scenarios is that the environments in which the robots are utilized have a high degree of certainty. Operating environments are often designed to be robot friendly, for example augmented reality markers are strategically placed and the ground is typically smooth, level, and clear of debris. For robots to be useful in a wider range of environments, especially environments that are not sanitized for their use, robots must be able to handle uncertainty. This requires a robot to incorporate new sensors and sources of information, and to be able to use this information to make decisions regarding navigation and the overall mission. When using autonomous mobile robots in unstructured and poorly defined environments, such as a natural disaster site or in a rural environment, ground condition is of critical importance and is a common cause of failure. Examples include loss of traction due to high levels of ground water, hidden cavities, or material boundary failures. To evaluate a non-contact sensing method to mitigate these risks, Frequency Modulated Continuous Wave (FMCW) radar is integrated with an Unmanned Ground Vehicle (UGV), representing a novel application of FMCW to detect new measurands for Robotic Autonomous Systems (RAS) navigation, informing on terrain integrity and adding to the state-of-the-art in sensing for optimized autonomous path planning. In this paper, the FMCW is first evaluated in a desktop setting to determine its performance in anticipated ground conditions. The FMCW is then fixed to a UGV and the sensor system is tested and validated in a representative environment containing regions with significant levels of ground water saturation.

Research funded by Offshore Robotics for the Certification of Assets Hub [EP/R026173/1] and Robotic and AI in NuclearHub [EP/R026084/1]

J. Blanche, S. Chirayil Nandakumar, D. Mitchell and S. Harper are with Heriot-Watt University, Edinburgh, UK, EH14 4AS (corresponding author e-mail: J.Blanche@hw.ac.uk).

K. Groves, A. West, B. Lennox and S. Watson are with The University of Manchester, UK, M13 9PL.

D. Flynn is with James Watt School of Engineering, University of Glasgow, UK, G12 8QQ

I. Yamamoto is with Nagasaki University, Nagasaki, 852-8521, Japan (e-mail:iyamamoto@nagasaki-u.ac.jp)

## I. INTRODUCTION

The use of robotics in the industrial and commercial sectors is well established, with facilities increasingly designed around the needs of a robotic fleet [1]. The predictable operating conditions within such facilities are optimized towards the efficient operation of wheeled robotic agents, for example smooth concrete flooring in warehouses. However, the practical application of robotic systems in dynamic environments of the real world requires run time path planning capable of identifying a safe route for navigation through areas less suitable for robotic operations [2]. Accounting for environmental dynamism in path planning for Robotic Autonomous Systems (RAS), while an established field, places a focus on object detection and collision avoidance as part mobile Simultaneous Location And Mapping (SLAM) operations. However, less research emphasis is placed on ground condition and ground integrity monitoring as a means of determining a route of safe passage through an area of uncertain ground integrity that may otherwise impede the RAS. In a warehouse environment, such ground integrity contrasts could be typified as fluid spillage on smooth concrete, representing compromised traction for a loaded wheeled robotic agent and a significant threat to autonomous control.

Uncertainty in outdoor ground conditions can be typified by stratigraphic layering due to depositional processes and with a high likelihood for interlaminar slippage due to material contrasts attributable to the differing stratigraphic layers e.g., snow on ice, or bedrock under sand. The resulting collapse of the ground condition represents a significant hazard to a robotic system and this research is aimed at developing an early warning sensor for unstable terrain detection, triggering dynamic path planning and autonomous decision-making for evasive action, where necessary. In natural and human built environments, terrain hazards exist in many forms, such as loose sand, fractured rocks or concrete, boggy soils, muds (Fig. 1A), snow and ice (Fig. 1B), all representing significant threats to mobility and mission continuity [3].

The state-of-the-art in edge dynamic path planning has resulted in autonomous mobile robots that can account for uncertainty in the conditions of its operational environment, with navigational decisions made based on information from a suite of sensors. Costmaps, gridded map representations of risk used in 2D path planning and navigation, can provide semi-autonomous systems with awareness of the position of static and mobile obstacles in an environment using sensors, such as lidar or depth cameras. Path planning algorithms then use this information to avoid areas where a collision may occur. This costmap approach provides robots with a spatially resolved representation of risk, which can be expanded to a range of sensing modalities and threats. An example is the integration of ionizing radiation sensing with Unmanned Ground Vehicles (UGVs) to map radiation levels in deployment areas where human presence is denied due to the radioactive environment. This places greater emphasis on edge sensing to both detect and map high risk areas, in addition to safeguarding the robotic platform by preventing a condition of irretrievable failure. The consequent evasion of highly localized radiation levels improves mission resilience and advances the RAS capacity for non-intervention in a hazardous environment [4]–[6].

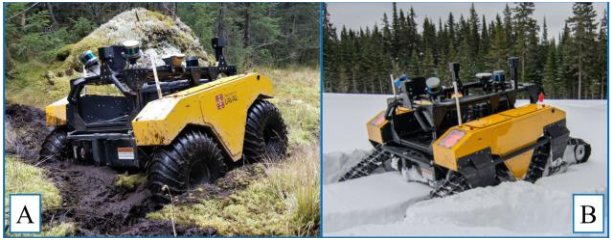

Figure 1. Clearpath Warthog during autonomous navigation research by the Northern Robotic Laboratory (Norlab), Québec. A) shows the UGV stranding in boggy conditions. B) shows a tracked conversion of the UGV sinking into deep snow [3].

FMCW sensing in the X- and K-bands has been successfully applied to the detection of contrasts in layered low to medium dielectric media, such as snow and ice stratigraphy, with sensors deployed on hauled sleds to generate a stratigraphic image of the snow subsurface [7]–[11]. In the K-band, FMCW has been found to be effective for contrast detection in soils [12], [13], fluid presence in sandstones, sands and concretes [14], in addition to the detection of failure precursors in loaded sandstones [15]. In wider applications, FMCW is an emergent sensing modality in many other sectors, such as medical monitoring [16], [17], automotive [18]–[22], aerospace [23]–[25], security and surveillance [26]–[28], energy sector asset integrity [29], [30].

A key enabler towards the robotic deployment of the above sensor modes includes the integration with RAS to inform their decision making and controlling algorithms [31], at the agent scale, while also providing condition parameters of the operating environment. This environmental data can be shared with a digital twin of the operating area, providing enhanced synchronicity and symbiosis between agents in a multi-robot fleet [32]–[35]. As such, this research represents the novel fusion of emergent sensor modes with the state-of-the-art in autonomous dynamic path planning for robotic agents required to operate in hazardous or Beyond Visual Line Of Sight (BVLOS) conditions. The successful integration of FMCW radar with autonomous environmental characterisation and mapping has the potential to provide new measurands of terrain integrity data, such as the detection of water, snow, ice, oil or other contaminants on the operating surface which may inhibit the operation/motion of a UGV. FMCW can also provide subsurface data if the surface is of low to medium dielectric strength. Providing robotic systems with situational awareness of ground integrity via FMCW sensing allows these systems to autonomously avoid associated hidden or previously undetectable navigation risks, therefore increasing their operational lifetime and reducing the need for intervention. Understanding the subsurface in real world operating environments will require consistent detection of contrasts and instabilities in stratigraphic layers, where the passage of a robot would have a high likelihood of slippage or stranding. This also translates into the potential for hidden object or void detection in snow, sand, soil, ice and some rock types.

This paper presents the benchmarking and testing phase of a K-band FMCW sensor system to detect surface moisture and standing water on concrete at the Field Robotics Laboratory at Heriot-Watt University, representing an analogue of a warehouse floor robotic operating area. The addition of moisture and standing water is representative of an operating hazard in an area optimized for robotic operations; where the presence of water is damaging to robot systems and a threat to traction for loaded autonomous ground vehicles. For example, this may be due to cargo spillage or warehouse roof failure.

The remainder of this paper is structured as follows; Section II presents the FMCW operating parameters and outlines FMCW operation. Section III presents the experimental tests and results. Section III.A presents static testing, where the FMCW sensor is held affixed above a concrete test area to give an unchanging field of view. This is to establish baselines and contrasts under simple operating conditions. Section III.B presents dynamic testing, where the FMCW is mounted on a Clearpath Husky A200 UGV, via a Universal Robotics UR5 manipulator arm, and concrete is scanned during transit of the robot over the test area. This is to evaluate the stability of the signal and suitability for integration with thresholding algorithms and subsequent use in a costmap for terrain mapping and autonomous path planning. Section IV discusses the acquired data and section V concludes.

II. EQUIPMENT SETUP AND OPERATING PARAMETERS

The K-band radar module was connected directly to a Flann 21-240 Standard Gain Horn (SGH) antenna via a SMA/SMA connector rated to 26 GHz, with key operating parameters given in Table 1 [36], [37]. Fig. 2 provides a block diagram of the sensor setup and analytical workflow. Evaluation of the utilized Flann Microwave antenna shows it to have a peak amplitude spot size on the target of ~36.4 millimetres radius at an antenna – target separation of 10 cm. Assuming a consistent near field divergence of 15º (Fig. 3A) for the radiation pattern emitted from this antenna, the Field Of View (FOV) for a target separation of 30 cm is ~20 x$10^{-3}$ m$^2$. Within this FOV, a minimal phase differential is observed (Figure 3B) [36], [37].

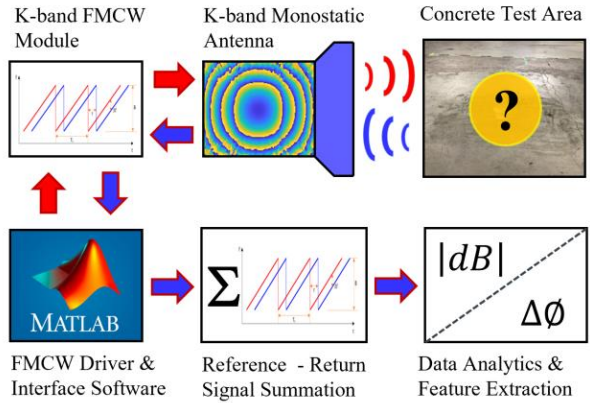

Figure 2. Block diagram of the sensor setup and analytical workflow, where the red (clockwise) arrows indicate the transmitted signal stages and the blue (counter-clockwise) arrows represent the received signal and data processing stages.

TABLE 1. FMCW PARAMETERS FOR STATIC AND DYNAMIC TRIALS

| Parameter | Value |
|---|---|
| Band | K- Band (24 – 25.5 GHz) Bandwidth 1500 MHz |
| Chirp Duration | 300 milliseconds |
| Field Of View on Target (30 cm) | 20 x $10^{-3}$ $m^2$ |
| Intermediate Frequency of Target (30 cm) | 12 MHz (or 24.012 GHz) |
| Sample Rate | 0.5 Hz |
| Data Transmit Time | 1.2 seconds |
| Analysis Time | 31 milliseconds |

### III. EXPERIMENTAL TESTS AND RESULTS

#### A. Static Testing

To evaluate the sensitivity of a statically-mounted FMCW sensor to moisture on smooth concrete, two experiments were conducted. Experiment A used an area of dry concrete as a baseline, prior to the passing of a wet cloth over the test area after a period of 10 seconds (Fig. 4A). Experiment B repeated Experiment A with the deposition of ~20 milliliters of water in the sensor FOV after ~10 seconds (Fig. 4B).

Fig. 5 shows the amplitude response of the FMCW sensor for each contrast agent condition within the test area. Clear contrasts are observed for both applied moisture conditions, experiments A and B, with a correlation between volume of water present and amplitude response. The same correlation is seen in Fig. 6, which shows the phase shift response of the FMCW sensor, and where an increase in water volume in the sensor FOV corresponds to an increasingly negative shift in the return signal phase.

#### B. Dynamic Testing: Motion on an Autonomous Vehicle

The Clearpath Husky A200 UGV-mounted FMCW can be seen in Fig. 7, where the antenna is protected within a low dielectric PolyLactic Acid (PLA) enclosure to allow a gripping point for the UR5 manipulator. The tip of the antenna is 30 cm from the concrete surface. The extent of the test area is indicated by the blue overlay and is bounded by tape markers on the concrete floor. The FOV of the sensor, indicated by the orange and yellow cone overlay, traverses an area where moisture was applied as a contrast agent and that is flanked by regions of dry concrete within the test area. Data was acquired under three surface moisture conditions:

1. A *"dry"* control scan, where the robot advances slowly over the test area, which was not wetted. This dataset acts as a baseline.

2. A *"damp"* scan, using the same test area boundaries and rate of transit as before, but with the midpoint of the test area dampened with a wet cloth.

3. A *"wet"* scan with the same test area saturated with ~100 ml of water.

In each instance, the movement of the Husky A200 is from point 1 to point 2, as indicated in Fig. 7.

Fig. 8 shows the Return Signal Amplitude (RSA) response at an Intermediate Frequency (IF) of 12 MHz for the three surface moisture conditions. This IF corresponds to a 30 cm height from the ground to the sensor tip. The legend in Fig. 8 describes the sequence of target conditions, with each start and end point being dry. A key observation of these datasets shows, as in the static data, there is a direct relationship between RSA and water volume in the sensor FOV. The relationship between the surface moisture conditions for the return signal phase can be seen in Fig. 9, where clear responses are evident. However, the phase response is similar for both moisture presence states, with the dry area exhibiting a more distinct contrast.

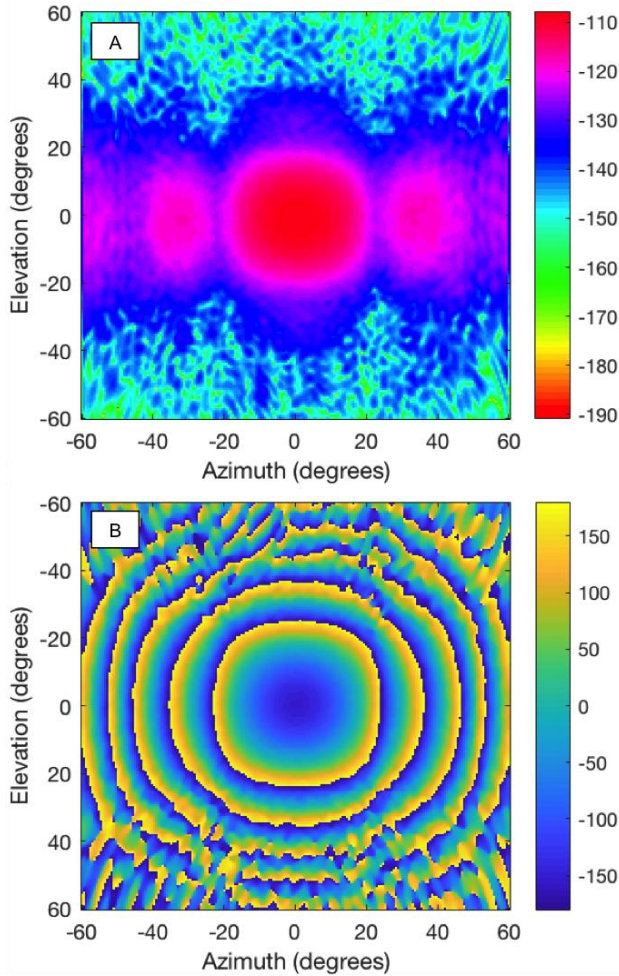

Figure 3. Emission characterization detail for K-Band Flann Microwave antenna model 21240-20. All data taken on a plane at 10 cm antenna – probe separation. A) Amplitude (scale bar in dBm) and B) Phase Shift (scale bar in degrees) [37].

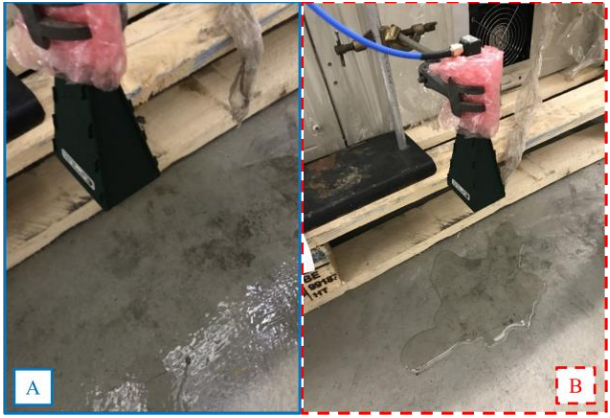

Figure 4. A) A wet cloth was applied to dampen the test area, note light surface sheen due to moisture. B) ~20 ml of water deposited in the sensor field of view.

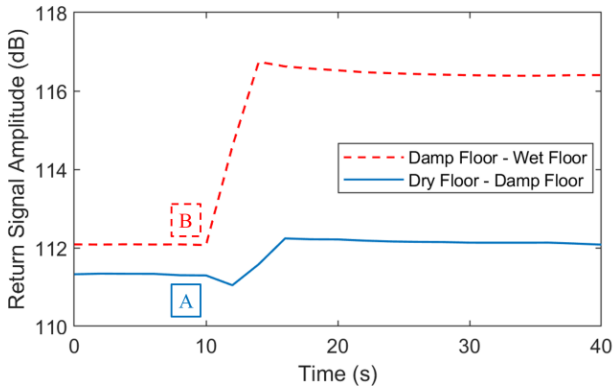

Figure 5. Static FMCW return amplitude over time for the intermediate frequency corresponding to the concrete floor interface at 12 MHz within the frequency sweep (or 24.012 GHz). A) Experiment A and B) Experiment B

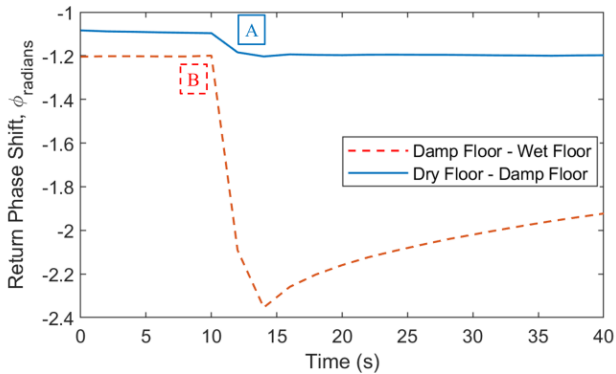

Figure 6. Static FMCW return phase shift over time for the intermediate frequency corresponding to the concrete floor interface at 12 MHz. A) Experiment A and B) Experiment B

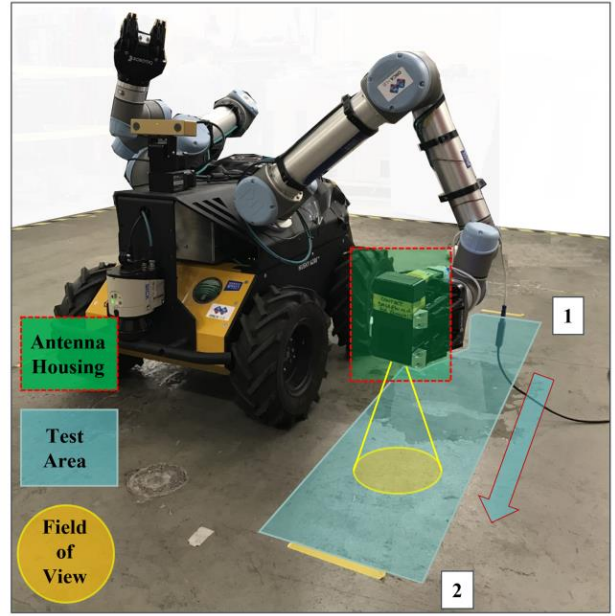

Figure 7. Experimental setup and cartoon overlay of test area and sensor field of view. Sensor tip is at a height of 30 cm. Points 1 and 2 indicate the direction of travel for each data acquisition

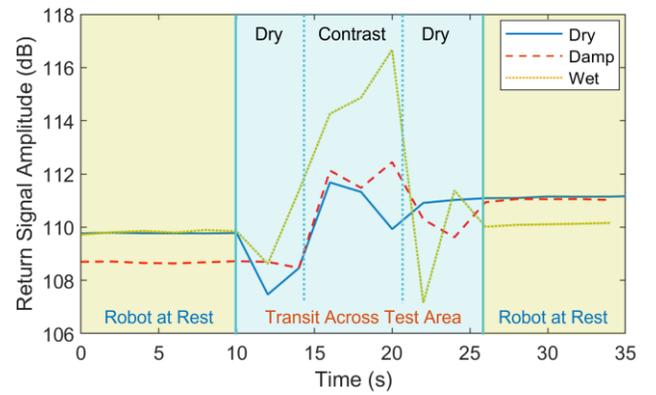

Figure 8. Dynamic FMCW return amplitude over time for the intermediate frequency corresponding to the concrete floor interface at 12 MHz.

## IV. Discussion

Static testing shows that the RSA is sensitive to the volume of water in the sensor FOV, with observed consistency between the dry concrete baseline and increasing degrees of water presence. This same consistency is also observed in the phase shift, with a higher degree of water in the sensor FOV resulting in a higher phase shift from a common initial condition. These tests validate the suitability of the FMCW sensor as a fixed position sensor to inform on ground condition contrasts within a single FOV.

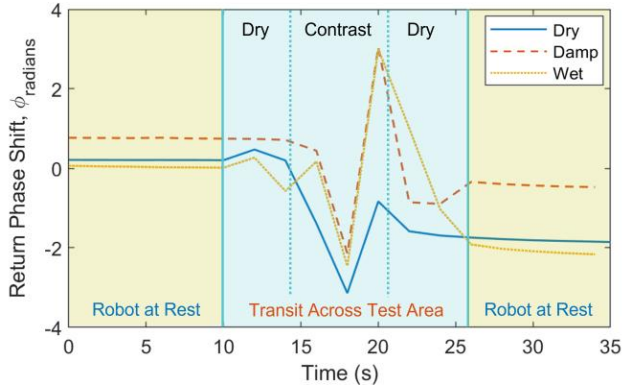

Figure 9. Dynamic FMCW return phase shift over time for the intermediate frequency corresponding to the concrete floor interface at 12 MHz.

Dynamic testing shows clear signal contrasts in varying surface moisture condition scenarios for both amplitude and phase responses. The relationships observed in the static testing phase are consistent with the dynamic data acquired, however, this phase of testing has identified the need to adapt the sensing parameters to improve the rate of data acquisition and signal variance. The advantages of an increased rate of data acquisition are twofold: to allow data averaging and thresholding algorithms, such as applied in [5], [6] to update a costmap closer to real time. Higher data acquisition rates also allow for faster UGV transit through the test area.

As discussed in [5], the sampling rate must be of sufficient frequency that the UGV can update its costmap, plan new paths to avoid areas of undesirable ground integrity, and the robot begin execution of the new path, all prior to committing to traversing an undesirable area. This practically results in sample rates faster than 1.0 Hz being necessary. Furthermore, interpolation will be required in the costmap to inflate FMCW observations into the configuration space of the robot. However, the fluctuations seen in dynamic testing in either RSA may become spatially averaged out, effectively masking the clear difference seen in the static tests. The consequence of this may be that a robot is unable to reliably distinguish between dry areas and those with only a small amount of moisture when in motion. Despite this, FMCW demonstrates itself as a strong candidate method to provide autonomous robotic systems with awareness of surface and subsurface ground integrity.

## V. CONCLUSION

The provision of autonomous path planning and run-time evasive action decision-making in the event of an identified terrain hazard, has significant cross-sector and cross-application potential. The FMCW sensing modality has been shown to be effective for the detection of surface contaminants and integration with a UGV has demonstrated that this novel application of millimetre wave sensing has the potential to add to full-field non-contact sensing capabilities in SLAM for RAS.

The future incorporation of robotically deployed FMCW sensitivity to these measurands will lead to enhanced situational awareness, improving the ability of field robots to operate in the dynamic and harsh environments of the real world with a minimum of human intervention. To achieve this, this research aims to evaluate the FMCW sensing modality on wheeled and quadruped robots operating in uneven and unprepared terrain, such as loose sand or deep snow, where data from the FMCW will feed into onboard robotic autonomous navigation software. Additions to the software will allow the robot to make decisions regarding path planning and mission feasibility based on this additional information stream.

ACKNOWLEDGMENTS

Industrial support from MicroSense Technologies Ltd (MTL) via use of equipment and facilities. Research funded by Offshore Robotics for the Certification of Assets (ORCA) Hub [EP/R026173/1], the Robotic and AI in Nuclear (RAIN) Hub [EP/R026084/1] and ALACANDRA consortium [EP/V026941/1].